\documentclass[conference]{IEEEtran}
\IEEEoverridecommandlockouts
\usepackage{cite}
\usepackage{amsmath,amssymb,amsfonts}
\usepackage{algorithmic}
\usepackage{graphicx}
\usepackage{textcomp}
\usepackage{xcolor}

\usepackage{booktabs}
\usepackage{caption}
\usepackage{makecell}
\usepackage{url} % 在导言区引入包
\newcommand{\Tref}[1]{Table~\ref{#1}}
\newcommand{\eref}[1]{Eq.~\eqref{#1}}

\newcommand{\fref}[1]{Fig.~\ref{#1}}

\newcommand{\Sref}[1]{Section~\ref{#1}}

\usepackage{xcolor}
\newcounter{todos}
\AtEndDocument{\ifnum\value{todos}>0 \PackageWarning{TODOS}{There are \arabic{todos} todos left in this paper! Fix them before submitting the paper!} \fi}

\usepackage{bm}

\usepackage{mathtools}

\usepackage{xspace}
\def\eg{\emph{e.g.}}

\def\BibTeX{{\rm B\kern-.05em{\sc i\kern-.025em b}\kern-.08em
    T\kern-.1667em\lower.7ex\hbox{E}\kern-.125emX}}

\begin{document}

\title{NRGS: Neural Regularization for Robust 3D Semantic Gaussian Splatting}

\author{
Zaiyan Yang$^{1,2\dagger}$, Xinpeng Liu$^{3\dagger}$, Heng Guo$^{1,2*}$, 
Jinglei Shi$^{4}$, Zhanyu Ma$^{1,2}$, Fumio Okura$^{3}$\\
$^{1}$Beijing University of Posts and Telecommunications, China\\
$^{2}$Beijing Key Laboratory of Multimodal Data Intelligent Perception and Governance, China\\
$^{3}$The University of Osaka, Japan\\ 
$^{4}$Nankai University, China\\
\texttt{\{738654937, guoheng, manzhanyu\}@bupt.edu.cn}\\
\texttt{\{liu.xinpeng, okura\}@ist.osaka-u.ac.jp};
\texttt{jinglei.shi@nankai.edu.cn}
}

% \maketitle

\newboolean{putfigfirst}
\setboolean{putfigfirst}{true}  % 设置为true，启用题图模式

\ifthenelse{\boolean{putfigfirst}}{
    \twocolumn[{%  % 注意：这里没有 %
        \renewcommand\twocolumn[1][]{#1}%
        \vspace{-0.3in}
        \maketitle\thispagestyle{empty}
        
        \begin{center}
            % \vspace{-0.3in}
            \includegraphics[width=\linewidth]{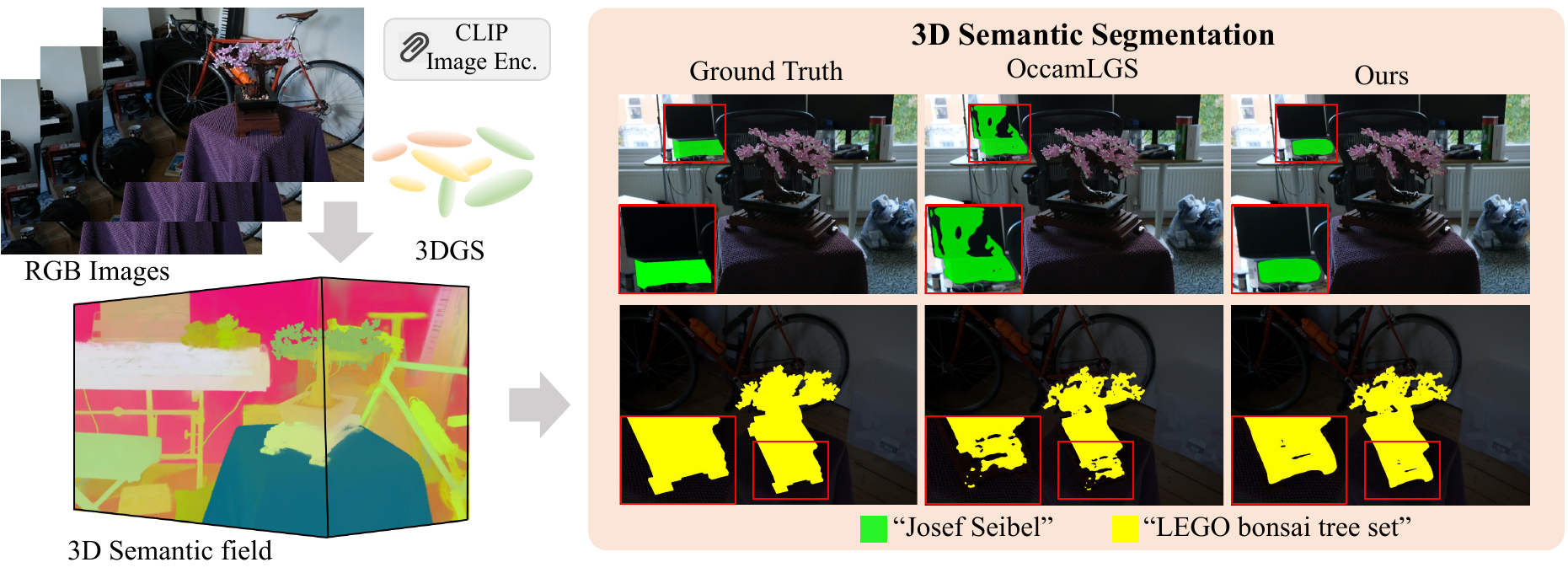}
            \vspace{-3mm}
            \captionof{figure}{3D semantic segmentation results on 3DGS point clouds. While the existing methods (\eg, Occam's LGS~\cite{occam}) do not consider consistency among multi-view 2D features, our NGRS corrects the inconsistency errors simultaneously with the optimization of 3D Gaussians, resulting in accurate segmentation.}
            \vspace{1mm}
            \label{fig:teaser}
        \end{center}%
    }]
}{
    \maketitle\thispagestyle{empty}
}

\begingroup
\renewcommand\thefootnote{}\footnote{$^\dagger$ Equal contribution. $^*$ Corresponding author.}
\addtocounter{footnote}{-1}
\endgroup

\begin{abstract}
 
We propose a neural regularization method that refines the noisy 3D semantic field produced by lifting multi-view inconsistent 2D features, in order to obtain an accurate and robust 3D semantic Gaussian Splatting.
The 2D features extracted from vision foundation models suffer from multi-view inconsistency due to a lack of cross-view constraints. Lifting these inconsistent features directly into 3D Gaussians results in a noisy semantic field, which degrades the performance of downstream tasks. 
Previous methods either focus on obtaining consistent multi-view features in the preprocessing stage or aim to mitigate noise through improved optimization strategies, often at the cost of increased preprocessing time or expensive computational overhead. 
In contrast, we introduce a variance-aware conditional MLP that operates directly on the 3D Gaussians, leveraging their geometric and appearance attributes to correct semantic errors in 3D space.  
Experiments on different datasets show that our method enhances the accuracy of lifted semantics, providing an efficient and effective approach to robust 3D semantic Gaussian Splatting. 
% Code is available at \url{https://anonymous.4open.science/r/NRGS-0969}. 

\end{abstract}


\section{Introduction}
\label{sec:intro}

% Intro部分参考PolGS的格式
% 第一段最后一句明确说明我们要输出一个什么东西，不要介绍其他方法做了什么
Understanding 3D scenes is a key task for applications like robot navigation~\cite{huang2022visual}, self-driving~\cite{zheng2024genad}, and augmented reality~\cite{izadi2011kinectfusion, klein2007parallel}. 
A major goal is to enable open-vocabulary
understanding, where a system can recognize objects based
on free-form language descriptions, not just a fixed list of
categories.
Accordingly, a desirable approach is to take 2D semantic features from vision-language models as input and output a semantic 3D field within the 3D Gaussian Splatting (3DGS)~\cite{3DGS} representation.

% 分别从两个挑战入手，第一个是如何把高维度的语义映射到3DGS，LangSplat和OccamLGS分别提出不同的解决方案。第二个是面临多视角语义不一致的问题，简单介绍2D多视角不一致是如何产生的，以及直接将其提升到3D可能产生什么危害 -- 然后引出我们的工作，凸显我们能够解决这些noisy的3D语义（其实还是需要稍微提一嘴GAGS，因为摘要中有所提及。下一段的开头需要表明To 快速且高效的 adress the noisy ... we...）；
A unique challenge is the incorporation of high-dimensional 2D semantic features into 3D space. LangSplat~\cite{langsplat} employs a scene-specific autoencoder for feature compression, then lifts the compressed features into 3D through optimization-based distillation; however, this can compromise training speed and semantic fidelity. In contrast, OccamLGS~\cite{occam} directly aggregates uncompressed 2D features into 3D via a weighted average scheme, preserving both speed and accuracy.
Another challenge remains the multi-view inconsistency of the extracted 2D features. Specifically, changes in viewpoint and occlusion can lead to significant semantic drift across different views of the same 3D region. When such inconsistent 2D features are lifted into 3D, they result in a noisy and unreliable semantic field. Several methods~\cite{GAGS, egosplat} have been proposed to address this problem. However, this often comes at the cost of either extensive pre-processing or high computational iterative optimization, limiting their practical applicability.

% 首先介绍我们的insight，然后介绍具体的method，这两个不要混合在一起
% Insight_v1：我们借助3DGS可靠的几何和外观属性，对lifted的语义进行正则化，消除了那些由于2D多视角不一致导致的语义错误。由于这个过程是直接在3DGS上进行优化的，因此是快速的。
% InsightP_v2：我们的观察是高斯的语义应当和几何和外观属性之间是存在联系，而非孤立，因此我们通过引入一个轻量级的MLP来建立高斯的基本属性和语义之间的映射。通过这种正则化手段，能够消除那些由于2D多视角不一致导致的语义错误。并且由于这个过程是直接在3DGS上进行优化的，因此是快速的。

% To refine the noisy semantic fields caused by multi-view inconsistent 2D features, we propose NRGS. 
To refine noisy semantic fields from multi-view inconsistencies, we propose NRGS. Our approach is motivated by the observation that the semantics of Gaussians should align with their intrinsic geometric and appearance attributes. We introduce a lightweight MLP to learn this mapping, thereby regularizing the semantic field. This process effectively corrects semantic errors. As optimization operates directly on the 3DGS representation, the method remains efficient.

% Method_v1：为了实现这一点，我们首先采用和OccamLGS一致的training-free策略来快速的将多粒度的2D特征进行lift。不同于OccamLGS，我们额外计算了每个lifted的3D高斯语义对应的方差，显式考虑它们的置信度。随后我们训练了一个轻量级的MLP来学习从3DGS几何外观到语义的连续映射，作为对高斯语义场的正则化。MLP能够根据几何和外观属性推断语义，而不是单纯的依赖多视角不一致的2D特征。
% 我们在训练过程中基于高斯语义的方差设置不同的损失权重，确保MLP能够进行鲁棒的学习。并且我们使用同一个MLP学习高斯属性到三个语义粒度的映射，这能使得MLP学习到更通用的表示，而不是在某个语义粒度上过拟合。

% Method_v2：这一段不要把主要重心放在别人的方法上，我们简单提一嘴OccamLGS即可，表明我们保持这种efficency。主要重心还是需要介绍方差的作用，以及我们的网络“支持”这种shared的方式来进行condition控制

Following the training-free lifting strategy~\cite{occam} for efficiency, we project 2D semantic features of multiple granularities into 3D. More importantly, we estimate the variance of each lifted semantic as a confidence measure for the noisy multi-view signals. Then we train a lightweight conditional MLP to learn a continuous mapping from the attributes of Gaussians to their corresponding semantics, acting as a direct regularizer on the Gaussian semantics. 
% However, treating all Gaussians equally during training is suboptimal. 
However, treating all Gaussians equally is suboptimal.
Therefore, we employ a variance-weighted training objective that adaptively prioritizes the learning from more reliable (lower-variance) Gaussians, ensuring robust regularization. Furthermore, by employing a single, shared MLP to refine semantics across multiple granularities, we encourage the network to learn a more general and unified representation rather than overfitting to any single semantic granularity. Collectively, our method builds a robust 3D semantic field, leading to improved performance as shown in~\fref{fig:teaser}.

% 此处的汇总不应该是第一次才出现哪些概念，应该是前面Intro已经介绍过的
% 
To summarize, the key contributions of this paper are:

\begin{itemize}
    \item We propose NRGS, a neural regularization method that refines noisy semantics for robust 3D semantic Gaussian Splatting.
    \item We introduce a variance-weighted training objective. It uses the variance of the initial lifted semantics as a confidence measure, enabling the model to adaptively focus on reliable signals for robust regularization.
    \item We condition a single, shared MLP on multiple semantic granularities, promoting the learning of a general and unified representation.

\end{itemize}

\section{Related works}
% In this section, we review the existing works on powerful 2D foundation models and the methods that integrate 2D features into 3D Gaussians, explaining their limitations and how our approach addresses these challenges.
In this section, we first review powerful 2D foundation models that provide the features essential for 3D understanding. We then discuss methods that lift these 2D features into 3D Gaussians, highlighting the key challenge of multi-view inconsistency and how our approach addresses this problem.

\vspace{-0.2cm}
\subsection{2D Open-Vocabulary Scene Understanding}
% Recent progress in 2D open-vocabulary scene understanding has been driven by the emergence of large-scale vision-language foundation models. CLIP~\cite{clip} provides a shared embedding space for images and text through contrastive pre-training, enabling open-vocabulary image classification and serving as a foundational semantic prior for dense prediction tasks. Building on this, prevailing methods such as LSeg~\cite{lseg} and SegCLIP~\cite{segclip} directly distill knowledge from CLIP~\cite{clip} to achieve open-vocabulary semantic segmentation. Concurrently, the Segment Anything Model (SAM)~\cite{SAM} has emerged as a notable foundational model for image segmentation. It offers powerful zero-shot instance segmentation, generating high-quality masks for a vast array of objects. Despite their distinct strengths, both CLIP-based and SAM-based methods are inherently designed for and limited to the 2D domain. When it comes to 3D scenes, the prediction result of these 2D models often suffer from severe inconsistencies across different viewpoints.

Recent progress in 2D open-vocabulary scene understanding has been driven by large-scale vision-language foundation models. CLIP~\cite{clip} learns a shared image-text embedding space via contrastive pre-training, enabling open-vocabulary classification and providing semantic priors for dense prediction. Building on this, methods such as LSeg~\cite{lseg} and SegCLIP~\cite{segclip} distill CLIP knowledge for open-vocabulary semantic segmentation. Meanwhile, SAM~\cite{SAM} enables high-quality zero-shot instance segmentation. However, both CLIP- and SAM-based approaches are limited to 2D, and their predictions often exhibit significant cross-view inconsistencies in 3D scenes.

\vspace{-0.2cm}
\subsection{3D Open-Vocabulary Scene Understanding}
3DGS~\cite{3DGS} has demonstrated strong performance in real-time novel view synthesis by representing scenes explicitly via anisotropic Gaussians. 
% To enable 3D open-vocabulary understanding, a line of research focuses on integrating semantic features from 2D foundation models into this 3D representation. 
To enable 3D open-vocabulary understanding, prior work integrates semantic features from 2D foundation models into this representation.
% Early efforts such as LangSplat~\cite{langsplat} and Feature 3DGS~\cite{feature3dgs}, propose integrating features from 2D foundation models using optimization-based feature distillation, where embeddings are lifted into 3D space through iterative optimization. LangSplatV2~\cite{langsplatv2} addresses LangSplat's critical inference bottleneck by introducing a sparse coefficient representation, achieving real-time 3D language querying. Subsequent approaches, including OccamLGS~\cite{occam} and LUDVIG~\cite{ludvig}, adopt training-free schemes, aggregating 2D features into 3D with a weighted average method rather than explicit backpropagation. 
Early methods such as LangSplat~\cite{langsplat} and Feature 3DGS~\cite{feature3dgs} rely on optimization-based feature distillation to lift 2D embeddings into 3D, while LangSplatV2~\cite{langsplatv2} improves efficiency via sparse coefficient representation for real-time querying. Later approaches, including OccamLGS~\cite{occam} and LUDVIG~\cite{ludvig}, adopt training-free schemes by aggregating 2D features through weighted averaging.

% A common limitation of the above methods is that they do not explicitly account for the multi-view inconsistency inherent in the 2D feature maps. The direct integration of these inconsistent features inevitably injects noise and ambiguity into the 3D semantic field. To address this, GAGS~\cite{GAGS} enhances multi-view consistency by linking SAM's prompt point density with camera distances and introducing an unsupervised granularity factor to selectively distill consistent 2D features. Similarly, EgoSplat~\cite{egosplat} proposes a multi-view consistent instance feature aggregation method, which eliminates semantic conflicts from occlusion through high-quality viewpoint selection and improves semantic feature representation via cross-view consistency aggregation. However, these methods either require extensive pre-processing to establish cross-view correspondence or rely on a time-consuming optimization process, limiting their flexibility and efficiency. 

A common limitation of these methods is that they overlook the multi-view inconsistency inherent in 2D feature maps, where directly integrating inconsistent features introduces noise and ambiguity into the 3D semantic field. To address this, GAGS~\cite{GAGS} improves multi-view consistency by relating SAM prompt point density to camera distance and introducing an unsupervised granularity factor for selectively distilling consistent features. Similarly, EgoSplat~\cite{egosplat} proposes a multi-view consistent instance feature aggregation strategy that mitigates occlusion-induced semantic conflicts through high-quality viewpoint selection and enhances representations via cross-view aggregation. However, these approaches either require extensive pre-processing to establish cross-view correspondence or rely on time-consuming optimization, limiting their efficiency and flexibility.

% In contrast, our work introduces a neural regularization method that regularizes the lifted 3D semantic field by establishing a mapping between the Gaussian attributes and corresponding semantics. This approach circumvents the need for complex pre-processing or costly optimization during lifting, offering a simpler yet effective pathway to robust 3D semantics.

In contrast, we introduce a neural regularization method that corrects the lifted 3D semantic field by learning a mapping from Gaussian attributes to semantics. This avoids complex pre-processing or costly optimization during lifting, offering a simpler yet effective path to robust 3D semantics.

\section{Method}
% \begin{figure}
% 	\includegraphics[width=\linewidth]{IEEE Conference Template - ICME 2026/images/pipe_ours.pdf}
% 	\caption{
%     % Our method imposes 3D consistency on semantically noisy Gaussians, enabling accurate 3D semantic understanding.
%     We propose a post-processing regularization method to construct a robust 3D semantic Gaussian Splatting.
%     }
% 	\label{fig:pipe}
% \end{figure}

\begin{figure*}
	\includegraphics[width=\linewidth]{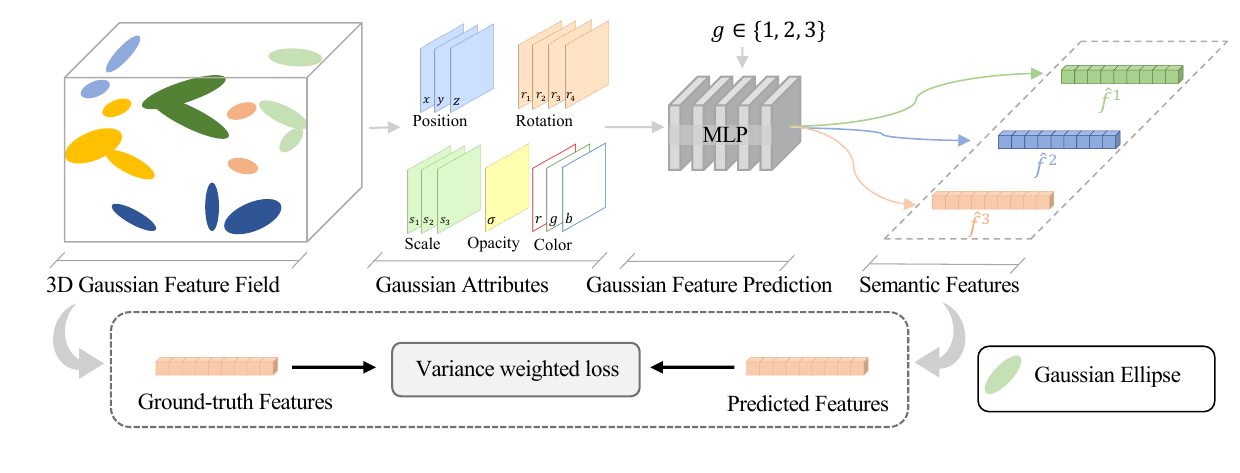}
    % \vspace{-1mm}
	\caption{Details of our neural regularization. A shared conditional MLP takes Gaussian attributes and a granularity condition $g$ as input, and is trained with a variance-weighted loss to regularize the noisy lifted semantics.}
	\label{fig:network}
\end{figure*}

% In this section, we present our variance-aware neural regularization method for 3D semantic Gaussians, with an overview in Fig.~\ref{fig:pipe}.
In this section, we present our neural regularization method for 3D semantic Gaussians. We first recall 3DGS~\cite{3DGS}, then describe how 2D semantic features are lifted into 3D without iterative training. Finally, we introduce our regularization method, detailing its network architecture and variance-weighted training objective for robust learning.

\vspace{-0.2cm}
\subsection{Preliminary}
A 3DGS~\cite{3DGS} scene is represented by a set of anisotropic 3D Gaussians $\mathcal{G} = \{G_i\}$. 
Each Gaussian $G_i$ is parameterized by its center position $\boldsymbol{\mu}_i$, covariance matrix $\boldsymbol{\Sigma}_i$ governing its shape and orientation, opacity $\sigma_i$, and spherical harmonics coefficients $\mathbf{c}_i$ for view-dependent appearance.
The 3D Gaussian function is defined as:
\begin{equation}
G_i(\mathbf{x}) = \exp\left(-\frac{1}{2}(\mathbf{x} - \boldsymbol{\mu}_i)^\top \boldsymbol{\Sigma}_i^{-1} (\mathbf{x} - \boldsymbol{\mu}_i)\right).
\end{equation}

To render an image, these 3D Gaussians are projected onto the 2D image plane. The color $\mathbf{C}$ at a pixel is computed via alpha blending, sorting the Gaussians by their depth:
\begin{equation}
\mathbf{C} = \sum_{i \in \mathbf{N}} \mathbf{c}_i \alpha_i \prod_{j=1}^{i-1} (1 - \alpha_j) = \sum_{i \in \mathbf{N}} \mathbf{c}_i \alpha_i T_i = \sum_{i \in \mathbf{N}} \mathbf{c}_i w_i,
\label{eq:rendering_weight}
\end{equation}
where $\mathbf{N}$ is the ordered list of Gaussians overlapping the pixel, $T_i = \prod_{j=1}^{i-1} (1 - \alpha_j)$ is the transmittance, and $w_i = \alpha_i T_i$ is the final weight of the $i$-th Gaussian. This differentiable rendering process is the foundation for optimizing the Gaussian parameters from multi-view images.

\vspace{-0.2cm}
\subsection{Semantic Feature Lifting}
\label{subsec:lifting}
While some prior methods~\cite{langsplat, feature3dgs} optimize the features during 3DGS training, this leads to a long training time. Following recent training-free feature aggregation technique~\cite{occam}, we adopt a more efficient and scalable lifting strategy.

Our goal is to assign a semantic feature vector \(\mathbf{f}_i \in \mathbb{R}^D\) to each 3D Gaussian \(G_i\) by leveraging 2D feature maps \(\{\mathbf{F}_m\}_{m=1}^M\) extracted from the input images. This is achieved by aggregating the 2D features from all views in which a Gaussian is visible, weighted by its contribution to each pixel.

For a Gaussian \(G_i\) visible in a set of source views \(\mathcal{S}_i\), we project its center to each view \(s \in \mathcal{S}_i\), obtaining the pixel coordinate \(\mathbf{u}_i^s\). The 2D feature \(\mathbf{F}_s(\mathbf{u}_i^s)\) at this location is retrieved. The contribution weight \(w_i^s\) of the Gaussian in view \(s\) is defined as its marginal rendering weight at the projected pixel,  which is given by~\eref{eq:rendering_weight}.

The initial semantic feature \(\mathbf{f}_i\) for Gaussian \(G_i\) is then computed as the weighted average:
\begin{equation}
\mathbf{f}_i \approx \frac{\sum_{s \in \mathcal{S}_i} w_i^s \cdot \mathbf{F}_s(\mathbf{u}_i^s)}{\sum_{s \in \mathcal{S}_i} w_i^s}.
\label{eq:feature_lifting}
\end{equation}

This direct aggregation scheme is efficient. However, the features obtained through~\eref{eq:feature_lifting} can inherit inconsistencies from the 2D predictions. To quantify this unreliability, we compute a per-Gaussian feature variance.

The variance of the approximated feature \(\mathbf{f}_i\) is estimated across the source views. Ignoring covariance between feature dimensions for efficiency, the variance for the \(d\)-th dimension is approximated following ~\cite{cf3} as:
\begin{equation}
\operatorname{Var}\left(\mathbf{f}_i\right)_d \approx \frac{\sum_{s \in \mathcal{S}_i} w_i^s \cdot \left( \mathbf{F}_s(\mathbf{u}_i^s)_d \right)^2 }{\sum_{s \in \mathcal{S}_i} w_i^s} - \left(\mathbf{f}_i\right)_d^2.
\label{eq:feature_variance}
\end{equation}
Gaussians with accurate geometry and consistent 2D feature projections across views will have low variance. In contrast, Gaussians with inaccurate positions or those on object boundaries will average conflicting information, resulting in high variance. This variance measure \(\operatorname{Var}(\mathbf{f}_i)\) serves as a crucial confidence metric for the initial lifted features.

\vspace{-0.2cm}
\subsection{Neural Regularization}
\label{subsec:regularization}

As mentioned in~\Sref{subsec:lifting}, the initial lifted features are noisy due to inconsistent 2D predictions. We observe that the multi-view consistent attributes of a 3D Gaussian provide a strong prior for inferring consistent semantics. To harness this prior, we propose a neural regularization approach.

Our key idea is to learn a continuous mapping from the consistent 3D attribute space to the semantic space using a lightweight conditional MLP. This mapping effectively corrects erroneous features caused by inconsistent multi-view observations. Crucially, this process acts as a direct and efficient regularizer, operating in minutes per scene without the need for expensive rendering of high-dimensional features.

\subsubsection{Network Architecture}

As illustrated in~\fref{fig:network}, our MLP is designed as a conditional model that takes two input streams: (1) the fundamental attributes of a Gaussian, including its position, opacity, rotation, scale, and RGB color; and (2) a scalar conditioning value $g \in \{1,2,3\}$ specifying the desired semantic granularity. This conditional design, enabling a single model to generate multi-granularity features, acts as an effective regularizer. It requires the model to maintain structural consistency across granularities, which encourages learning a more generalized representation and mitigates overfitting to noisy labels at any single level.

% This conditional design, which enables a single model to generate multi-granularity features, serves as an effective regularizer. By requiring the shared backbone to simultaneously produce coherent outputs for subpart, part, and whole semantics, it discourages the network from overfitting to the noise in any single granularity's labels. This encourages the learning of a more generalized and unified representation, as the model must capture the underlying hierarchical structure of the scene to satisfy all constraints at once.

The architecture employs a residual structure to facilitate stable gradient flow during training. The final output is the regularized 512-dimensional semantic feature $\hat{\mathbf{f}}_i^g$ for the Gaussian at the specified granularity $g$.

\subsubsection{Variance-Weighted Optimization Objective}

A straightforward approach to train the MLP is to employ a loss function that equally weights the contribution of all Gaussians. For a batch of $N$ Gaussians at the same granularity, where $\mathbf{f}_i$ is the initial lifted feature of the $i$-th Gaussian and $\hat{\mathbf{f}}_i$ is the MLP's refined prediction, such an equal-weighted loss can be formulated as: % 修改：y_i -> f_i, \hat{y}_i -> \hat{f}_i
\begin{equation}
\mathcal{L}_{\text{equal}} = \frac{1}{N} \sum_{i=1}^{N} \left[ \|\mathbf{f}_i - \hat{\mathbf{f}}_i\|_2^2 + \lambda_{\text{cos}} \cdot \left(1 - \frac{\mathbf{f}_i \cdot \hat{\mathbf{f}}_i}{\|\mathbf{f}_i\|_2 \|\hat{\mathbf{f}}_i\|_2} \right) \right], % 修改：y_i -> f_i, \hat{y}_i -> \hat{f}_i
\label{eq:loss_equal}
\end{equation}
which combines a Mean Squared Error (MSE) term with a cosine similarity term, balanced by a coefficient $\lambda_{\text{cos}}$.

However, this naive weighting strategy is suboptimal. The initial features $\mathbf{f}_i$ are not equally reliable. Gaussians with accurate geometry and consistent multi-view projections yield low-variance features, which serve as high-quality supervision. In contrast, Gaussians with inaccurate positions or those residing on semantic boundaries inherit conflicting 2D predictions, resulting in high-variance, noisy features. Treating all Gaussians as equally important allows these unreliable, high-variance points to pollute the learning process of the MLP.

To address this, we introduce a variance-weighted loss function that adjusts the importance of each Gaussian during training based on its confidence. We leverage the per-Gaussian feature variance $\operatorname{Var}(\mathbf{f}_i)$ computed in~\eref{eq:feature_variance} as a confidence measure. The variance is first condensed into a scalar uncertainty $v_i = \|\operatorname{Var}(\mathbf{f}_i)\|_2$, which is then normalized and mapped to a weight $p_i$ that is inversely proportional to the variance. This mapping uses a temperature parameter $\gamma$ to control the sensitivity of the weighting: 
\begin{equation}
p_i = \exp\left(-\gamma \cdot \tilde{v}_i\right), \quad \text{with} \quad \tilde{v}_i = \frac{v_i - \min(\mathbf{v})}{\max(\mathbf{v}) - \min(\mathbf{v}) + \epsilon}.
\label{eq:variance_weight}
\end{equation}
The final, variance-aware optimization objective is then defined as:
\begin{equation}
\mathcal{L} = \frac{1}{N} \sum_{i=1}^{N} p_i \cdot \left[ \|\mathbf{f}_i - \hat{\mathbf{f}}_i\|_2^2 + \lambda_{\text{cos}} \cdot \left(1 - \frac{\mathbf{f}_i \cdot \hat{\mathbf{f}}_i}{\|\mathbf{f}_i\|_2 \|\hat{\mathbf{f}}_i\|_2} \right) \right]. % 修改：y_i -> f_i, \hat{y}_i -> \hat{f}_i
\label{eq:final_loss}
\end{equation}
This variance-weighted scheme effectively acts as an adaptive filter, allowing the training objective to focus on regions of the 3D scene where the initial lifted features are most accurate, thereby enhancing the robustness of the regularization process.

\section{Experiments}
We compare our method to state-of-the-art methods on open vocabulary segmentation and localization tasks, along with their computational efficiency.

% \vspace{-0.2cm}
\subsection{Datasets and Details}
\subsubsection{Datasets}
We evaluate our method on the LERF~\cite{lerf} and Mip-NeRF-360~\cite{mip-nerf} datasets. The LERF dataset~\cite{lerf}, containing in-the-wild scenes captured by a smartphone, is used for open-vocabulary 3D object localization and segmentation based on LangSplat's protocol~\cite{langsplat}. For 3D segmentation, we further employ the Mip-NeRF-360 dataset~\cite{mip-nerf}, leveraging its complex indoor and outdoor scenes and the ground-truth annotations from GAGS~\cite{GAGS}.

\subsubsection{Baselines}
We compare our method with recent relevant works including LangSplat~\cite{langsplat}, GAGS~\cite{GAGS}, LangSplatV2~\cite{langsplatv2} and OccamLGS~\cite{occam}. We employ GSplat~\cite{gsplat} to construct initial Gaussian fields. For a fair comparison, all methods are evaluated under the same setting.

% All comparative methods are evaluated under identical setting to ensure a fair comparison.

\subsubsection{Evaluation metrics}
We adopt the evaluation protocol from LangSplat~\cite{langsplat} to assess performance on text-based 3D localization and segmentation. For localization, we report the mean accuracy (mAcc). For segmentation, we report the mean Intersection-over-Union (mIoU).

\subsubsection{Implementation Details}
Following LangSplat~\cite{langsplat}, we utilize SAM ViT-H~\cite{SAM} and OpenCLIP ViT-B/16~\cite{clip} for segmentation and CLIP feature extraction. To hierarchically capture semantic information, we generate 2D feature maps at three distinct levels of granularity: whole, part, and subpart. The 3D Gaussians are first trained with RGB supervision for 30,000 iterations to reconstruct the scene. Subsequently, we lift the 2D semantic features into the 3D scene using the training views exclusively. A regularization process is then applied to the semantic features in the 3D space. The final evaluation is performed on held-out testing views and all experimental results reported in this paper were re-evaluated. For a fair comparison, we evaluate GAGS~\cite{GAGS} using the standard LangSplat~\cite{langsplat} pipeline features, replacing its original granularity-aware sampling input. Finally, all experiments are conducted on a single RTX 4090 GPU.

% \vspace{-0.2cm}
\subsection{Comparison with Baselines Methods}

\subsubsection{Open Vocabulary Semantic Segmentation}

\begin{figure*}
    \centering
	\includegraphics[width=\linewidth]{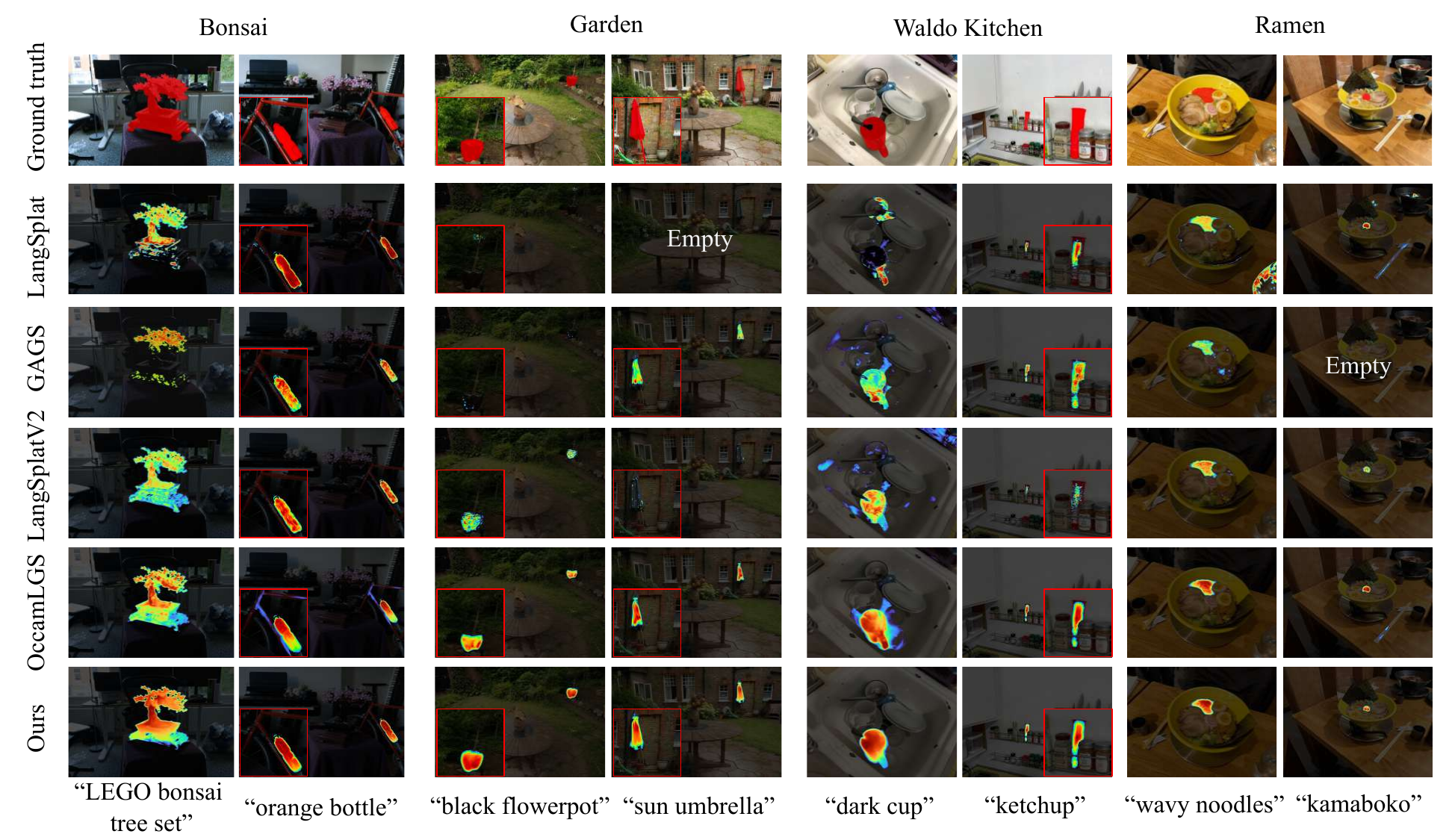}
	% \caption{Qualitative comparison of relevancy score visualization.}
	\caption{Visualization of the relevance score for open-vocabulary queries, with the ground-truth segmentation of the target object shown in red.}
    \label{fig:Qualitative}
\end{figure*}

\begin{table}[htbp]
    \setlength\tabcolsep{2pt}
    \centering
    \caption{Comparison of average IoU on LERF Dataset.}
    \label{tab:lerf_comparison}
    \begin{tabular}{lccccc}
        \toprule
        \textbf{Method}       & \textbf{Ramen} & \textbf{Figurines} & \textbf{Teatime} & \textbf{\makecell{Waldo \\ Kitchen}} & \textbf{Overall} \\
        \midrule
        LangSplat~\cite{langsplat}& 46.8& 43.3& 48.6& 38.0& 44.2\\
        GAGS~\cite{GAGS}& 42.1& 52.3& 54.9& 56.9& 51.6\\
        LangSplatV2~\cite{langsplatv2}& 48.7& \textbf{59.8}& 65.1& 61.1& 58.7\\
        OccamLGS~\cite{occam}& \underline{50.2}& \textbf{59.8}& \underline{72.6}& \underline{65.3}& \underline{62.0}\\
        \textbf{Ours}        &\textbf{52.0}& \underline{57.8}& \textbf{73.5}& \textbf{65.8}& \textbf{62.3}\\
        \bottomrule
    \end{tabular}
\end{table}

\begin{table}[htbp]
    \setlength\tabcolsep{3pt}
    \centering
    \caption{Comparison of average IoU on Mip-NeRF-360 Dataset.}
    \label{tab:mip_comparison}
    \begin{tabular}{lccccc}
        \toprule
        \textbf{Method}       & \textbf{Room}& \textbf{Counter} & \textbf{Garden} & \textbf{Bonsai} & \textbf{Overall} \\
        \midrule
        LangSplat~\cite{langsplat}& 49.6& 71.3& 52.4& 49.5& 55.7\\
        GAGS~\cite{GAGS}& 67.1& 59.2& 57.6& 58.0& 60.5\\
        LangSplatV2~\cite{langsplatv2}& 61.0& \textbf{75.5}& 56.6& 59.3& 63.1\\
        OccamLGS~\cite{occam}& \underline{68.2}& 73.6& \textbf{60.9}& \underline{59.4}& \underline{65.5}\\
        \textbf{Ours}        &\textbf{69.0}& \underline{75.4}& \underline{60.2}& \textbf{63.9}& \textbf{67.1}\\
        \bottomrule
    \end{tabular}
\end{table}

The quantitative results in~\Tref{tab:lerf_comparison} and~\Tref{tab:mip_comparison} and the qualitative results shown in~\fref{fig:Qualitative} demonstrate the superior performance of our method on both the LERF~\cite{lerf} and Mip-NeRF-360~\cite{mip-nerf} datasets. This improvement across diverse scenes demonstrates the effectiveness of our approach. While OccamLGS~\cite{occam} and LangSplatV2~\cite{langsplatv2} also achieve competitive scores, their performance drop with complex occlusions because they do not explicitly resolve multi-view inconsistencies, yielding sub-optimal results. GAGS~\cite{GAGS} attempts to address this problem by selecting the most consistent granularity features during optimization, but this can cause a loss of semantic detail in complex areas. In contrast, our method leverages the intrinsic 3D consistency of the Gaussian attributes to regularize the semantic field. This enables our approach to mitigate the impact of multi-view inconsistent 2D features while retaining all multi-granularity information, thus yielding more accurate and robust 3D semantics.

\subsubsection{Localization}
We also evaluate the language-guided localization task on the LERF dataset~\cite{lerf} following LangSplat~\cite{langsplat}. As shown in~\Tref{tab:lerf_comparison_loc}, our approach achieves the highest localization accuracy of $86.6\%$, demonstrating its capability to construct a more accurate semantic field than existing approaches.

\begin{table}[htbp]
    \setlength\tabcolsep{3pt}
    \centering
    \caption{Comparison of localization accuracy on LERF Dataset.}
    \label{tab:lerf_comparison_loc}
    \begin{tabular}{lccccc}
        \toprule
        \textbf{Method}       & \textbf{Ramen} & \textbf{Figurines} & \textbf{Teatime} & \textbf{\makecell{Waldo \\ Kitchen}} & \textbf{Overall} \\
        \midrule
        LangSplat~\cite{langsplat}& 71.8& 75.0& 76.3& 68.2& 72.8\\
        GAGS~\cite{GAGS}& 63.4& 75.0& 86.4& \underline{77.3}& 75.5\\
        LangSplatV2~\cite{langsplatv2}& 71.8& \underline{82.1}& 88.1& 72.7& 78.7\\
        OccamLGS~\cite{occam}& \underline{73.2}& \underline{82.1}& \underline{91.5}& \underline{77.3}& \underline{81.0}\\
        \textbf{Ours}        &\textbf{74.7}& \textbf{87.5}& \textbf{93.2}& \textbf{90.9}& \textbf{86.6}\\
        \bottomrule
    \end{tabular}
\end{table}

\subsubsection{Runtime analysis}
\begin{figure}
	\includegraphics[width=\linewidth]{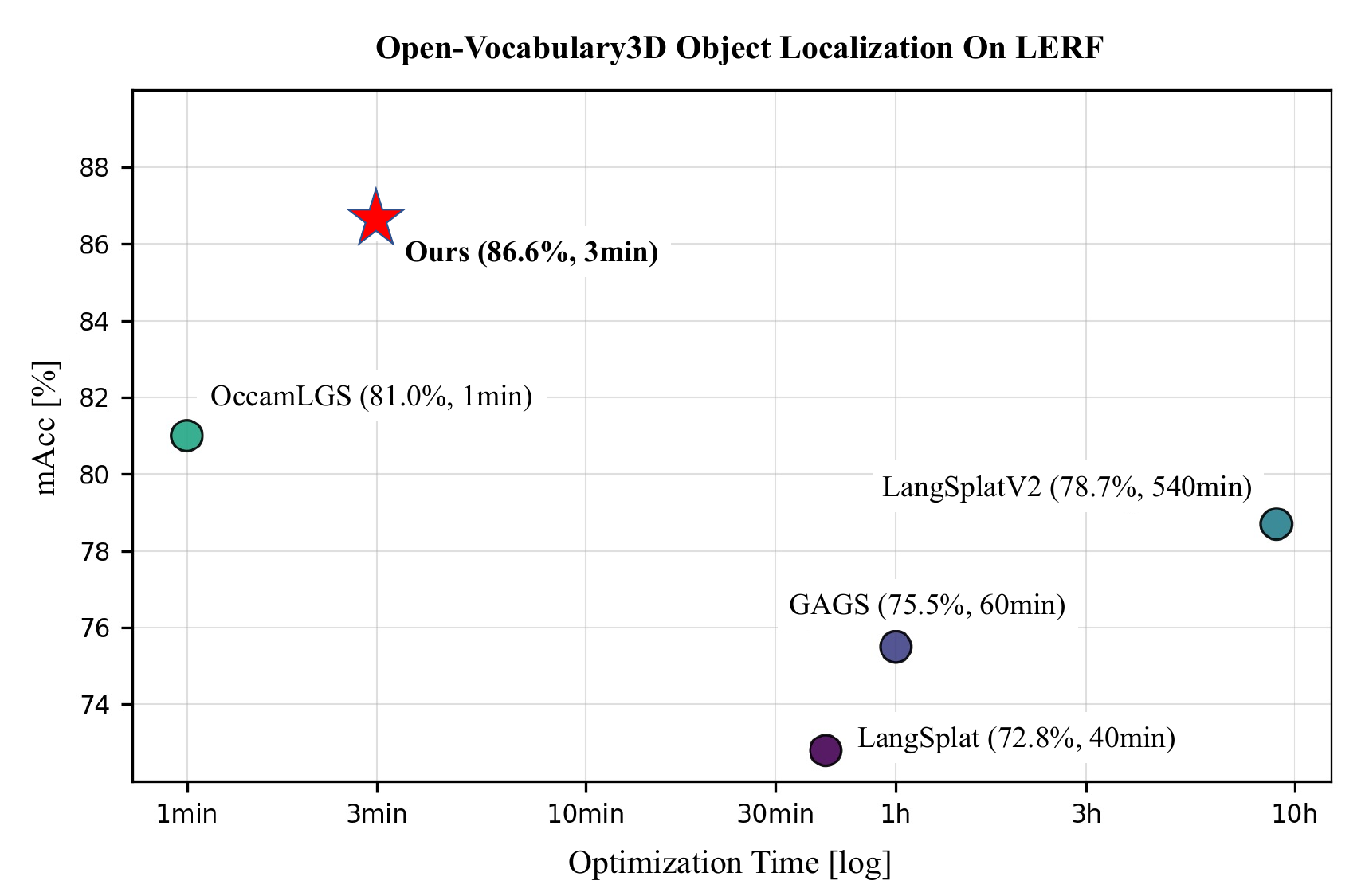}
	\caption{
    % Our method only takes a few minutes to achieve the most accurate performance. Comparison in 3D open-vocabulary localization on the LERF dataset~\cite{lerf}.
    Comparison in 3D open-vocabulary localization on the LERF dataset~\cite{lerf}. Our method only takes a few minutes to achieve the most accurate performance. 
    }
	\label{fig:time_analysis}
\end{figure}

We report the runtime of our method and baselines in~\fref{fig:time_analysis}. Although training-free method OccamLGS~\cite{occam} achieves the fastest optimization speed, its effectiveness can be affected when the input features contain significant noise. In contrast, optimization-based methods~(e.g., LangSplat~\cite{langsplat}, LangSplatV2~\cite{langsplatv2}, GAGS~\cite{GAGS}) incur substantial training overhead. Our method first efficiently lifts features into 3D via fast aggregation, then performs direct regularization on 3D space, which requires no expensive iterative training. The entire process completes in only a few minutes per scene, demonstrating its practical efficiency.

% \vspace{-0.2cm}
\subsection{Ablation Study}

As shown in~\Tref{tab:ablation_study}, we conduct an ablation study on the LERF dataset~\cite{lerf}, averaging the metrics over all scenes. Both proposed components, the variance-weighted loss and the multi-granularity conditioning, contribute positively to the final performance.  The model employing only the multi-granularity conditioned MLP already shows a clear improvement over the noisy baseline, demonstrating its effectiveness in regularizing the lifted features. Integrating the variance-weighted loss further boosts the performance, validating its role in robustly guiding the training process. Crucially, replacing the multi-granularity conditioning with independent MLPs per granularity causes a performance drop, as the shared representation is necessary for the variance-weighted loss to effectively resolve ambiguities in high-variance regions.

\begin{table}[htbp]
    \centering
    \caption{Ablation Studies on the LERF dataset.}
    \label{tab:ablation_study} 
    \begin{tabular}{c|cc|cc}
        \hline
        \# & Variance-Weighted & Multi-Granularity & mIoU $\uparrow$ & mAcc $\uparrow$ \\
        \hline
        a) & $\checkmark$ & & 60.7& 80.7\\
        b) & & $\checkmark$ & 61.8& 83.3\\
        c) & $\checkmark$ & $\checkmark$ & \textbf{62.3}& \textbf{86.6}\\
        \hline
    \end{tabular}
\end{table}

\section{Conclusion}
In this paper, we proposed a neural regularization method to refine noisy 3D semantic Gaussians. Our key insight is to leverage the intrinsic consistency of Gaussian attributes to regularize the unreliable 3D semantics derived from inconsistent 2D predictions. This is achieved through a lightweight conditional MLP, optimized with a variance-weighted objective that prioritizes learning from reliable multi-view signals. Experiments demonstrate that our method effectively enhances 3D semantic consistency and accuracy, offering an effective solution for robust 3D semantic understanding.

\section{Acknowledgment}
This work was supported by Beijing Major Science and Technology Project under Contract No. Z251100007125021, Hebei Natural Science Foundation Project No. 242Q0101Z, Beijing-Tianjin-Hebei Basic Research Funding Program No. F2024502017, National Natural Science Foundation of China (Grant No. 62472044, U24B20155), in part by the Beijing Key Laboratory of Multimodal Data Intelligent Perception and Governance.

\bibliographystyle{IEEEtran}
\bibliography{icme2026references}

\end{document}